\documentclass{article}
\usepackage{fontspec}

\usepackage{arxiv}

\usepackage{hyperref}       
\usepackage{url}            
\usepackage{booktabs}       
\usepackage{amsfonts}       
\usepackage{nicefrac}       
\usepackage{microtype}      
\usepackage{lipsum}		
\usepackage{graphicx}
\usepackage{doi}
\usepackage{algorithm}
\usepackage{algpseudocode}
\usepackage{float}
\usepackage{tabularx}
\usepackage{array}
\usepackage{authblk}

\newfontfamily\tamilfont{NotoSansTamilLight}[
    Path = ./fonts/,
    Script = Tamil,
    Language = Tamil,
    Extension = .ttf
]

\title{A Grapheme-Aware Indic Tokenizer for Tamil: Large-Scale Training and Intrinsic Evaluation}

\author[1]{Harikrishnan K V}
\author[2]{Sudarsun Santhiappan}
\affil[1]{Department of Computer Science and Engineering, IIITDM Kanchipuram, \texttt{hk.hari9700@gmail.com}}
\affil[2]{Wadhwani School of Data Science and AI, IIT Madras, \texttt{sudarsun@dsai.iitm.ac.in}}

\renewcommand{\shorttitle}{Grapheme-Aware Indic Tokenizer for Tamil}

\hypersetup{
pdftitle={A Grapheme-Aware Indic Tokenizer for Tamil: Large-Scale Training and Intrinsic Evaluation},
pdfsubject={Natural Language Processing},
pdfauthor={Harikrishnan K V, Sudarsun Santhiappan},
pdfkeywords={Tamil, Tokenization, Grapheme-aware Tokenization, WordPiece, Indic NLP, Intrinsic Evaluation}
}

\begin{document}

\maketitle


\begin{abstract}

Tokenization forms the foundation of modern Natural Language Processing (NLP) systems by transforming raw text into discrete units that neural language models can process. The effectiveness of this process directly influences vocabulary efficiency, sequence length, computational cost, and downstream model performance. Although multilingual tokenizers such as Byte Pair Encoding (BPE), WordPiece, and SentencePiece have performed well across numerous languages, they often segment morphologically rich Indic languages inefficiently. Tamil, in particular, poses unique challenges because its grapheme-based writing system can represent a single visible character with multiple Unicode code points. Conventional Unicode-level tokenizers often fragment these grapheme clusters into multiple subword units, increasing sequence lengths and reducing vocabulary efficiency.

In this work, we present a grapheme-aware Indic tokenizer for Tamil that preserves complete grapheme clusters through a reversible Unicode mapping strategy prior to WordPiece vocabulary learning. By operating on grapheme-level representations instead of individual Unicode code points, the tokenizer produces linguistically meaningful token boundaries while remaining fully compatible with transformer-based language models. The tokenizer is trained on a large-scale Tamil corpus and evaluated using a comprehensive intrinsic evaluation framework that measures compression efficiency, token fragmentation, information density, and vocabulary utilization.

Experimental evaluation compares the proposed tokenizer against five widely used multilingual tokenizers: GPT-2, mBERT, mT5, mBART, and NLLB. The proposed tokenizer achieves the strongest performance among the evaluated tokenizers on fragmentation- and sequence-efficiency-oriented intrinsic metrics, while matching the highest observed compression ratio. These results demonstrate the effectiveness of grapheme-aware preprocessing for Tamil tokenization.

\end{abstract}


\keywords{Tamil NLP \and Grapheme-aware Tokenization \and WordPiece \and Indic Languages \and Intrinsic Evaluation}


\section{Introduction}

Tokenization is one of the most fundamental stages of modern Natural Language Processing (NLP) pipelines. Every transformer-based language model first converts raw text into discrete tokens before learning contextual representations. Consequently, tokenization quality directly influences sequence length, vocabulary efficiency, computational requirements, and ultimately the effectiveness of downstream language models.

Modern multilingual language models predominantly employ subword tokenization algorithms such as Byte Pair Encoding (BPE), WordPiece, or SentencePiece. These approaches successfully balance vocabulary size with the ability to represent unseen words and have become the standard choice for models including GPT, mBERT, mT5, mBART, and NLLB. However, because multilingual vocabularies are jointly learned across hundreds of languages, vocabulary capacity must be shared among scripts exhibiting vastly different linguistic and morphological characteristics. This often results in suboptimal segmentation for low-resource and morphologically rich languages, including many Indic languages.

Tamil presents additional challenges due to its writing system. Unlike Latin-based scripts, Tamil text is naturally organised around grapheme clusters rather than individual Unicode code points. A single visible Tamil character may be composed of multiple Unicode code points representing consonants, vowels, and combining marks. Conventional tokenizers typically operate directly on Unicode sequences without explicitly preserving grapheme boundaries, frequently splitting complete grapheme clusters into multiple subword fragments. Such fragmentation increases token sequence lengths, reduces compression efficiency, and may adversely affect representation learning in downstream models.

Language-specific tokenization strategies provide an opportunity to address these limitations by incorporating script-level linguistic information during vocabulary construction. Grapheme-aware tokenization preserves complete grapheme clusters before subword learning, allowing vocabulary construction to operate on linguistically meaningful units while maintaining compatibility with existing transformer architectures. Such approaches are particularly attractive for Indic languages, where grapheme integrity closely reflects the underlying orthographic structure.

Evaluating tokenizer quality remains equally important. Most prior work measures tokenizer performance indirectly through downstream NLP tasks, making it difficult to isolate tokenizer behavior from model architecture and training procedures. Intrinsic evaluation offers a complementary perspective by directly quantifying tokenizer characteristics such as compression efficiency, fragmentation, sequence length, information density, and vocabulary utilization without requiring expensive model training.

In this work, we investigate a grapheme-aware Indic tokenizer for Tamil and evaluate its effectiveness using a comprehensive intrinsic evaluation framework. We train the tokenizer on a large-scale Tamil corpus and systematically compare it against widely used multilingual tokenizers. We employ multiple intrinsic metrics to analyze compression behavior, fragmentation characteristics, vocabulary efficiency, and token sequence quality independently of downstream language models.

The primary contributions of this work are summarised below.

\begin{itemize}

\item We present a grapheme-aware WordPiece tokenizer for Tamil that preserves complete grapheme clusters through reversible Unicode mapping.

\item We train the tokenizer on a large-scale Tamil corpus and investigate the impact of vocabulary scaling on tokenizer quality.

\item We establish a comprehensive intrinsic evaluation framework comprising compression, fragmentation, and vocabulary-based metrics for systematic tokenizer comparison.

\item We compare the proposed tokenizer with widely used multilingual tokenizers including GPT-2, mBERT, mT5, mBART, and NLLB.

\item Experimental results demonstrate that grapheme-aware tokenization substantially reduces fragmentation while maintaining competitive compression efficiency and vocabulary utilisation.

\end{itemize}


\section{Related Work}

Subword tokenization has become the dominant approach for neural language modelling because it effectively balances vocabulary size with the ability to represent previously unseen words. Among the earliest and most influential methods is Byte Pair Encoding (BPE), which iteratively merges frequently occurring symbol pairs to construct increasingly expressive vocabularies. WordPiece extends this concept by selecting merges according to probabilistic criteria, while SentencePiece removes language-dependent pre-tokenization entirely by learning segmentation directly from raw text.

These algorithms underpin many contemporary multilingual language models. GPT-2 employs byte-level BPE to achieve complete Unicode coverage, whereas multilingual BERT adopts WordPiece trained on multilingual Wikipedia corpora. More recent multilingual encoder-decoder models such as mT5, mBART, and NLLB utilise SentencePiece vocabularies learned from substantially larger multilingual datasets.

Despite their success, multilingual tokenizers often exhibit reduced efficiency for Indic languages because a shared vocabulary must simultaneously represent hundreds of languages. Consequently, relatively fewer vocabulary entries are allocated to individual Indic scripts, leading to increased token fragmentation, longer token sequences, and lower compression efficiency.

Tamil presents additional challenges because its orthography is fundamentally organised around grapheme clusters rather than individual Unicode code points. Conventional Unicode-level tokenizers generally ignore grapheme boundaries during vocabulary learning, frequently decomposing complete grapheme clusters into multiple subword fragments. Such segmentation reduces linguistic consistency and increases sequence length.

Recent research has therefore explored language-aware tokenization strategies for Indic languages, including morphology-aware tokenization, grapheme-aware segmentation, and script-specific vocabulary construction. Among these approaches, grapheme-aware tokenization offers a particularly attractive balance between linguistic preservation and compatibility with existing transformer architectures by treating complete grapheme clusters as atomic units prior to subword learning.

While numerous studies evaluate multilingual tokenizers using downstream NLP tasks, comparatively fewer investigations have focused on intrinsic tokenizer evaluation. Intrinsic metrics directly quantify tokenizer behavior independently of downstream language models, enabling systematic comparison of vocabulary efficiency, fragmentation, information density, and sequence compression. In this work, we adopt a comprehensive intrinsic evaluation framework to analyze grapheme-aware tokenization for Tamil and compare it with widely used multilingual tokenizers.


\section{Grapheme-Aware Indic Tokenizer}
\label{sec:tokenizer}
\subsection{Motivation}

Most existing multilingual tokenizers process text as sequences of Unicode code points and subsequently learn subword vocabularies using algorithms such as Byte Pair Encoding (BPE), WordPiece or SentencePiece. While this strategy performs well for many alphabetic writing systems, it is less suitable for Indic scripts where a single visible character is frequently composed of multiple Unicode code points.

Tamil follows an abugida writing system in which consonants, vowels and combining marks together form grapheme clusters. Treating these constituent code points independently often causes conventional tokenizers to split linguistically meaningful graphemes into multiple fragments. This fragmentation increases sequence length and reduces the information represented by individual tokens.

The proposed tokenizer addresses this issue through a grapheme-aware preprocessing stage that converts complete grapheme clusters into reversible atomic symbols before WordPiece vocabulary learning. Consequently, vocabulary construction operates on linguistically meaningful units while remaining fully compatible with standard transformer tokenization pipelines.


\subsection{Tokenizer Architecture}

The tokenizer consists of three major stages:

\begin{enumerate}

\item Grapheme extraction from normalized Tamil Unicode text.

\item Reversible mapping of each grapheme cluster to a unique Unicode representation.

\item WordPiece vocabulary learning over the mapped grapheme sequence.

\end{enumerate}

During inference, the reverse mapping restores the original Tamil Unicode representation after tokenization. Since the mapping is bijective, no linguistic information is lost during encoding or decoding.

\begin{figure}[!htbp]
\centering
\includegraphics[width=0.45\linewidth]{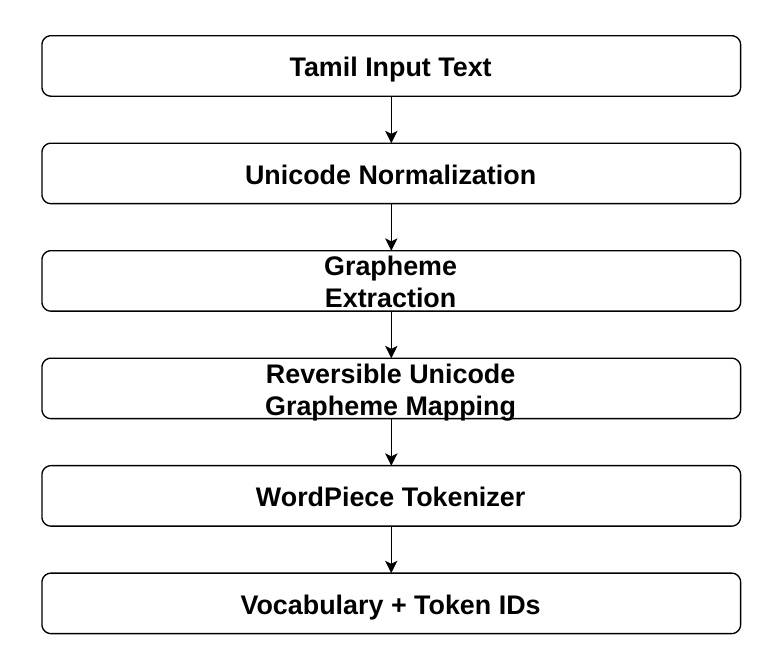}
\caption{Architecture of the proposed Grapheme-Aware Indic Tokenizer. The training pipeline performs Unicode normalization, grapheme extraction, reversible grapheme mapping and WordPiece vocabulary learning to construct a grapheme-aware vocabulary. During inference, the reverse mapping restores the original Tamil Unicode representation.}
\label{fig:architecture}
\end{figure}

Figure~\ref{fig:architecture} illustrates the overall tokenizer pipeline.

\subsection{Grapheme Mapping}

The central idea of the proposed tokenizer is to preserve complete Tamil grapheme clusters during vocabulary learning.

Rather than applying WordPiece directly on raw Unicode code-point sequences, each grapheme cluster is first mapped to a dedicated reversible Unicode representation. Consequently, every grapheme behaves as an indivisible atomic symbol during tokenizer training.

This mapping offers several advantages.

\begin{itemize}

\item Grapheme integrity is preserved throughout vocabulary construction.

\item Frequently occurring grapheme combinations can be learned directly as subword units.

\item The mapping is deterministic and fully reversible.

\item Existing WordPiece implementations can be used without modification.

\end{itemize}

Unlike script-specific rule-based tokenization, the proposed approach remains compatible with standard transformer tokenizers while incorporating script-aware linguistic information through preprocessing.


\subsection{WordPiece Vocabulary Learning}

Following grapheme mapping, the transformed corpus is used to train a WordPiece tokenizer.

WordPiece iteratively constructs a vocabulary by identifying frequently occurring symbol sequences and incorporating them as vocabulary entries. Since the input symbols correspond to complete grapheme clusters rather than individual Unicode code points, the learned vocabulary naturally captures larger Tamil subword units while preserving grapheme boundaries.

The tokenizer architecture remains unchanged throughout all experiments. Only the vocabulary size is varied in order to analyze the influence of vocabulary capacity on intrinsic tokenization quality.


\begin{algorithm}[H]
\caption{Training Procedure for the Grapheme-Aware Indic Tokenizer}
\label{alg:training}

\begin{algorithmic}[1]

\Require Tamil corpus $C$, vocabulary size $V$

\State Normalize Unicode text

\State Extract grapheme clusters

\State Map graphemes to reversible Unicode symbols

\State Train WordPiece tokenizer with vocabulary size $V$

\State Save vocabulary and reverse mapping

\Return Grapheme-aware tokenizer

\end{algorithmic}

\end{algorithm}


\section{Training Methodology}

The proposed Grapheme-Aware Indic Tokenizer was trained on a cleaned large-scale Tamil Wikipedia corpus. Prior to tokenizer training, the corpus was normalized and converted into grapheme-level representations using the reversible Unicode mapping strategy described in Section \ref{sec:tokenizer}.
We then used the transformed corpus to train WordPiece vocabularies of different sizes.

Table~\ref{tab:corpus_statistics} summarizes the characteristics of the training corpus after preprocessing.

\begin{table}[H]
\centering
\caption{Statistics of the cleaned Tamil Wikipedia training corpus.}
\label{tab:corpus_statistics}

\begin{tabular}{lr}
\toprule
Statistic & Value\\
\midrule
Wikipedia Articles & 569,055\\
Text Lines & 3,471,054\\
Total Words & 42,834,262\\
Unique Words & 4,383,638\\
Total Characters & 386,120,438\\
Average Words per Line & 12.34\\
Longest Line & 19,010 characters\\
Tamil Character Ratio & 75.4\%\\
Other Character Ratio & 24.6\%\\
\bottomrule
\end{tabular}

\end{table}

The resulting corpus is substantially larger than the corpus used during the initial development phase, increasing from approximately 50 thousand text lines and 0.79 million words to over 3.47 million text lines and 42.8 million words. This approximately 44-fold increase in sentence count and 54-fold increase in word count provides significantly richer lexical diversity for learning grapheme-aware subword representations. Furthermore, the corpus remains strongly Tamil-centric, with approximately 75.4\% of all characters belonging to the Tamil Unicode block while naturally preserving multilingual content such as English words, numerals and punctuation.

Table~\ref{tab:training_setup} summarizes the training configuration.

\begin{table}[H]
\centering
\caption{Training Configuration}
\label{tab:training_setup}

\begin{tabular}{ll}
\toprule
Parameter & Value\\
\midrule
Language & Tamil\\
Training Corpus & Tamil Wikipedia\\
Tokenizer Algorithm & WordPiece\\
Preprocessing & Grapheme-aware Unicode Mapping\\
Vocabulary Sizes & 10K, 20K, 30K, 50K\\
Training Strategy & Independent training for each vocabulary\\
Evaluation Framework & Intrinsic Evaluation\\
\bottomrule
\end{tabular}

\end{table}

Only the vocabulary size was varied during training, while the corpus, preprocessing pipeline, tokenizer architecture, training configuration and evaluation dataset remained unchanged. Consequently, any observed differences in tokenizer performance can be attributed solely to vocabulary scaling.


\subsection{Vocabulary Scaling}

To investigate the relationship between vocabulary size and tokenizer quality, four tokenizer models were trained using progressively larger vocabularies:

\begin{center}
\begin{tabular}{cc}
\toprule
Model & Vocabulary Size \\
\midrule
Tokenizer-10K & 10,000\\
Tokenizer-20K & 20,000\\
Tokenizer-30K & 30,000\\
Tokenizer-50K & 50,000\\
\bottomrule
\end{tabular}
\end{center}

Each tokenizer was evaluated independently using the same intrinsic evaluation framework.

By maintaining identical experimental conditions across all runs, the study
isolates the effect of vocabulary size on tokenizer performance. This
enables a systematic analysis of how vocabulary expansion influences
compression efficiency, token fragmentation, sequence length, and
vocabulary utilization without introducing confounding factors.


\section{Experimental Setup}

\subsection{Experimental Configuration}

We applied the intrinsic evaluation framework to every tokenizer under identical experimental conditions. We evaluated the proposed tokenizer and all baseline multilingual tokenizers using the same Tamil evaluation corpus to ensure a fair comparison.

Table~\ref{tab:exp_setup} summarizes the experimental configuration adopted throughout this work.

\begin{table}[H]
\centering
\caption{Experimental Configuration}
\label{tab:exp_setup}

\begin{tabular}{ll}
\toprule
Parameter & Value\\
\midrule
Training Language & Tamil\\
Training Corpus & Tamil Wikipedia\\
Evaluation Corpus & FLORES+ Tamil\\
Evaluation Samples & 5,000\\
Tokenizer Algorithm & WordPiece\\
Vocabulary Sizes & 10K, 20K, 30K, 50K\\
Baseline Tokenizers & GPT-2, mBERT, mT5, mBART, NLLB\\
Evaluation Method & Intrinsic Evaluation\\
Metrics & CR$_{max}$, Fertility, CPT, WFR, PPC, NSL\\
\bottomrule
\end{tabular}

\end{table}

All tokenizer models were evaluated on the same 5,000-sample Tamil subset
of FLORES+, using identical metric definitions and evaluation procedures.
This controlled setup reduces variation arising from dataset differences
and enables direct comparison of tokenizer behaviour.


\subsection{Baseline Tokenizers}

We compared the proposed Grapheme-Aware Indic Tokenizer against five widely used multilingual tokenizers representing different subword tokenization approaches.

\begin{itemize}

\item GPT-2
\item mBERT
\item mT5
\item mBART
\item NLLB

\end{itemize}

These tokenizers were selected because they are widely adopted multilingual models and provide strong baselines for evaluating Tamil tokenization quality.


\subsection{Intrinsic Evaluation Metrics}

We evaluated the proposed tokenizer using a comprehensive intrinsic evaluation framework.

\begin{itemize}

\item \textbf{Maximum Compression Ratio (CR$_{max}$):}
The maximum achievable compression ratio obtained immediately after the pretokenization stage. Higher values indicate more efficient compression of the input text before subword segmentation.

\item \textbf{Fertility:} Average number of generated tokens per pretoken. Lower values indicate less fragmentation.

\item \textbf{Characters per Token (CPT):} Average number of characters represented by each token. Higher values indicate more informative token representations.

\item \textbf{Word Fragmentation Rate (WFR):} Fraction of words segmented into multiple subword units. Lower values indicate better preservation of complete words.

\item \textbf{Percentage of Pretokens Changed (PPC):} Fraction of pretokens requiring additional segmentation. Lower values indicate more stable tokenization.

\item \textbf{Normalized Sequence Length (NSL):} Token sequence length normalized relative to GPT-2. Lower values correspond to shorter sequences.

\item \textbf{Vocabulary Coverage:} Percentage of pre-tokens directly represented within the tokenizer vocabulary.

\end{itemize}

These complementary metrics evaluate compression efficiency, segmentation
behavior, vocabulary utilization, information density, and sequence
length without requiring downstream language-model training.

Under the predominantly whitespace-based pre-tokenization strategy used
in the evaluation framework, pretoken boundaries closely correspond to
word boundaries. Consequently, WFR and PPC measure closely related
phenomena: WFR captures whether words are fragmented, whereas PPC captures
whether pretokens undergo additional segmentation. Their identical values
for the Grapheme-Aware IndicTokenizer therefore reflect the structure of
the current pre-tokenization strategy rather than an independent
mathematical equivalence between the two metrics.

\section{Results}

\subsection{Vocabulary Scaling}

Table~\ref{tab:vocab_scaling} presents the intrinsic evaluation results obtained after training the Grapheme-Aware Indic Tokenizer with progressively larger vocabularies.

\begin{table}[H]
\centering
\caption{Vocabulary Scaling Results}
\label{tab:vocab_scaling}

\begin{tabular}{lccccccc}

\toprule

Vocabulary &
CR &
Fertility &
CPT &
WFR &
PPC &
NSL &
Coverage \\

\midrule

10K &
9.0571 &
1.9841 &
4.5649 &
0.5769 &
0.5769 &
0.0806 &
99.9554 \\

20K &
9.0571 &
1.7562 &
5.1572 &
0.4859 &
0.4859 &
0.0713 &
99.9554 \\

30K &
9.0571 &
1.6566 &
5.4674 &
0.4396 &
0.4396 &
0.0673 &
99.9554 \\

50K &
9.0571 &
1.5527 &
5.8332 &
0.3884 &
0.3884 &
0.0631 &
99.9554 \\

\bottomrule

\end{tabular}

\end{table}


\subsection{Analysis of Vocabulary Scaling}

The experimental results demonstrate a consistent improvement in intrinsic tokenization quality as the vocabulary size increases.

Compression Ratio remains constant across all vocabulary sizes, indicating that overall corpus-level compression is largely unaffected by vocabulary expansion.

In contrast, Fertility decreases monotonically from 1.98 for the 10K tokenizer to 1.55 for the 50K tokenizer. This indicates that larger vocabularies enable the tokenizer to represent words using fewer subword units.

Characters per Token exhibits the opposite trend, increasing from 4.56 to 5.83 as vocabulary size increases. Consequently, each token carries progressively more linguistic information.

Word Fragmentation Rate and Percentage of Pretokens Changed both decrease steadily with increasing vocabulary size, demonstrating that larger vocabularies preserve complete words more effectively while requiring fewer additional segmentations.

Similarly, the Normalized Sequence Length decreases from 0.0806 to 0.0631, indicating that larger vocabularies produce shorter token sequences that may improve computational efficiency during downstream language-model training.

Vocabulary Coverage remains effectively constant at approximately 99.96\%
across all vocabulary sizes, suggesting that vocabulary expansion primarily
improves segmentation quality rather than lexical coverage. This indicates
that the additional vocabulary capacity is primarily used to learn longer
and more efficient subword units rather than to substantially increase
lexical coverage. 


\subsection{Comparison with Existing Multilingual Tokenizers}

We then compared the best-performing tokenizer (50K vocabulary) against widely used multilingual tokenizers.

\begin{table*}[t]

\centering

\caption{Comparison with Existing Multilingual Tokenizers}

\label{tab:comparison}

\begin{tabular}{lcccccccc}

\toprule

Tokenizer &
CR &

Fertility &

CPT &
WFR &
PPC &
NSL \\

\midrule

GPT-2 &
1.3576 &

24.6169 &

0.3679 &
0.9849 &
0.9937 &
1.0000 \\

mBERT &
7.5238 &

3.5913 &

2.5220 &
0.7974 &
0.6335 &
0.1459 \\

mT5 &
9.0571 &

2.4990 &

3.6243 &
0.8100 &
0.8100 &
0.1015 \\

mBART &
9.0008 &

2.4226 &

3.7385 &
0.6619 &
0.6578 &
0.0984 \\

NLLB &
9.0008 &

2.5415 &

3.5638 &
0.7085 &
0.7041 &
0.1032 \\

IndicTokenizer (50K) &
9.0571 &

1.5527 &

5.8332 &
0.3884 &
0.3884 &
0.0631 \\

\bottomrule

\end{tabular}

\end{table*}



\subsection{Qualitative Tokenization Analysis}
\label{sec:qualitative}

To complement the quantitative intrinsic evaluation, a qualitative
tokenization analysis was performed using a representative Tamil
sentence. The best-performing 50K Grapheme-Aware IndicTokenizer was
compared with the five multilingual baseline tokenizers used throughout
the evaluation: GPT-2, mBERT, mT5, mBART, and NLLB.

The input sentence used for the comparison was:

\begin{center}
{\tamilfont
\textbf{தமிழ் மொழி உலகின் மிகைப் பழமையான செம்மொழிகளில் ஒன்றாகும்.}
}
\end{center}

Table~\ref{tab:qualitative_tokenization} presents the tokenization
produced by each tokenizer for the same Tamil sentence.

\begin{table*}[t]
\centering
\caption{Qualitative tokenization comparison for the representative
Tamil sentence. The complete token sequences produced by the
50K Grapheme-Aware IndicTokenizer and the multilingual baseline
tokenizers are shown.}
\label{tab:qualitative_tokenization}

\renewcommand{\arraystretch}{1.25}

\begin{tabularx}{\textwidth}{
>{\bfseries\raggedright\arraybackslash}p{0.15\textwidth}
>{\raggedright\arraybackslash}X
>{\centering\arraybackslash}p{0.08\textwidth}
}

\toprule
Tokenizer & Tokenization & Token Count \\
\midrule

IndicTokenizer (50K) &
{\tamilfont
தமிழ் $\mid$ மொழி $\mid$ உலகின் $\mid$ மிகை $\mid$ \#\#ப்
$\mid$ பழமையான $\mid$ செம்மொழி $\mid$ \#\#களில்
$\mid$ ஒன்றாகும் $\mid$ .
}
& \textbf{10}
\\

\midrule

GPT-2 &

byte level fragments
& 157
\\

\midrule

mBERT &
{\tamilfont
தமிழ் $\mid$ மொழி $\mid$ உலக $\mid$ \#\#ின் $\mid$ மிக
$\mid$ \#\#ை $\mid$ \#\#ப் $\mid$ ப $\mid$ \#\#ழ
$\mid$ \#\#மையான $\mid$ ச $\mid$ \#\#ெ $\mid$ \#\#ம்
$\mid$ \#\#ம $\mid$ \#\#ொ $\mid$ \#\#ழி $\mid$ \#\#களில்
$\mid$ ஒன்றாகும் $\mid$ .
}
& 19
\\

\midrule

mT5 &
{\tamilfont
▁தமிழ் $\mid$ ▁மொழி $\mid$ ▁உலக $\mid$ ின் $\mid$ ▁மிக
$\mid$ ைப் $\mid$ ▁பழ $\mid$ மையான $\mid$ ▁செ $\mid$ ம்
$\mid$ மொழி $\mid$ களில் $\mid$ ▁ஒன்ற $\mid$ ாகும் $\mid$ .
}
& 15
\\

\midrule

mBART &
{\tamilfont
▁தமிழ் $\mid$ ▁மொழி $\mid$ ▁ $\mid$ உலகின் $\mid$ ▁மிக
$\mid$ ைப் $\mid$ ▁பழ $\mid$ மையான $\mid$ ▁செ $\mid$ ம்
$\mid$ மொழி $\mid$ களில் $\mid$ ▁ஒன்ற $\mid$ ாக $\mid$ ும்
$\mid$ .
}
& 16
\\

\midrule

NLLB &
{\tamilfont
▁தமிழ் $\mid$ ▁மொழி $\mid$ ▁உலக $\mid$ ின் $\mid$ ▁மிக
$\mid$ ைப் $\mid$ ▁பழ $\mid$ மையான $\mid$ ▁செ $\mid$ ம்ம
$\mid$ ொ $\mid$ ழி $\mid$ களில் $\mid$ ▁ஒன்ற $\mid$ ாகும்
$\mid$ .
}
& 16
\\

\bottomrule

\end{tabularx}

\end{table*}

The qualitative comparison reveals substantial differences in the
segmentation behaviour of the evaluated tokenizers. The 50K
Grapheme-Aware IndicTokenizer produces only 10 tokens for the complete
sentence and preserves several complete Tamil lexical units, including
``{\tamilfont
தமிழ்}'', ``{\tamilfont
மொழி}'', ``{\tamilfont
உலகின்}'', ``{\tamilfont
பழமையான}'', ``{\tamilfont
செம்மொழி}'', and
``{\tamilfont
ஒன்றாகும்}''. Only selected words are further segmented into subword
units, such as ``{\tamilfont
மிகைப்}'' into ``{\tamilfont
மிகை}'' and ``{\tamilfont
ப்}'', and
``{\tamilfont
செம்மொழிகளில்}'' into ``{\tamilfont
செம்மொழி}'' and ``{\tamilfont
களில்}''.

In comparison, mBERT produces 19 tokens and exhibits substantially finer
segmentation of several Tamil words. For example, ``{\tamilfont
மிகைப்}'' is
segmented into multiple subword units, while ``{\tamilfont
செம்மொழிகளில்}'' is
represented using several smaller components. The SentencePiece-based
tokenizers, mT5, mBART, and NLLB, preserve larger units than mBERT but
still fragment several Tamil words into smaller pieces.

GPT-2 exhibits the most severe fragmentation in this example, producing
157 tokens for the same sentence. Its byte-level representation results
in a very large number of byte-level token fragments compared with the
other tokenizers. Consequently, the same Tamil sentence requires more
than fifteen times as many tokens with GPT-2 as with the 50K
Grapheme-Aware IndicTokenizer.

The qualitative results are consistent with the quantitative comparison
presented in Table~\ref{tab:comparison}. The 50K IndicTokenizer produces
the shortest sequence for this example while preserving larger Tamil
subword units. This behavior is consistent with its lower Fertility and
Normalized Sequence Length and its higher Characters per Token observed
in the intrinsic evaluation.

Overall, the example provides a direct qualitative illustration of the
effect of grapheme-aware vocabulary construction. By preserving
grapheme clusters during vocabulary learning, the tokenizer can learn
larger and more linguistically meaningful Tamil subword units, thereby
reducing unnecessary token fragmentation while retaining compatibility
with the WordPiece framework.


\subsection{Discussion}

The comparative evaluation demonstrates that the proposed Grapheme-Aware Indic Tokenizer consistently achieves the strongest intrinsic performance for Tamil.

The tokenizer obtains the lowest Fertility, lowest Word Fragmentation Rate, lowest Percentage of Pretokens Changed, lowest Normalized Sequence Length and highest Characters per Token among all evaluated models. Despite using a language-specific vocabulary, it also matches the highest observed Compression Ratio.

These results indicate that grapheme-aware preprocessing enables WordPiece to learn substantially longer and more informative subword units while preserving Tamil grapheme boundaries. Consequently, the tokenizer generates shorter token sequences with considerably less fragmentation than existing multilingual tokenizers.

Overall, the experiments demonstrate that combining grapheme-aware
preprocessing with large-scale corpus training and vocabulary scaling
improves intrinsic tokenization quality for Tamil. The consistent
improvements in Fertility, CPT, WFR, PPC, and NSL, together with the
comparative benchmark results, provide evidence that grapheme-aware
preprocessing improves segmentation efficiency while preserving complete
grapheme units.


\section{Limitations}

Although the proposed Grapheme-Aware Indic Tokenizer demonstrates strong intrinsic performance for Tamil, several limitations remain.

First, the current study focuses exclusively on Tamil. While the tokenizer architecture is designed to support Indic scripts more broadly, its effectiveness on other languages such as Malayalam, Hindi, Telugu, Kannada and Bengali has not yet been evaluated.

Second, this work primarily relies on intrinsic evaluation metrics. Although intrinsic evaluation provides valuable insight into segmentation quality, vocabulary utilization, and token efficiency, it does not directly measure tokenization's influence on downstream Natural Language Processing tasks such as language modeling, machine translation, or question answering.

Third, the comparative evaluation considers widely used multilingual tokenizers but does not include recently proposed Indic-specific tokenizers. Incorporating such models would provide a more comprehensive understanding of the strengths and limitations of grapheme-aware tokenization within the broader Indic NLP ecosystem.

Finally, the tokenizer was trained using a Tamil Wikipedia corpus. Although this corpus provides broad linguistic coverage and high-quality text, it does not capture all domains of Tamil usage such as conversational language, social media, literary text or domain-specific corpora. Training on larger and more diverse corpora may further improve tokenizer robustness.


\section{Future Work}

Several directions remain for future investigation.

The most immediate extension of this work is multilingual evaluation across additional Indic languages. Since many Indic scripts share similar orthographic characteristics, it will be valuable to determine whether the benefits of grapheme-aware tokenization generalise beyond Tamil.

Future work should also investigate the impact of grapheme-aware tokenization on downstream transformer models. Integrating the tokenizer into encoder-only and encoder-decoder architectures will enable direct measurement of improvements in language modelling, machine translation, summarization and other NLP tasks.

Another promising direction is comparison with recently developed Indic-specific tokenizers and multilingual language models trained specifically for Indian languages. Such comparisons would provide a more complete assessment of the proposed tokenizer within the rapidly evolving Indic NLP landscape.

Finally, alternative vocabulary-learning algorithms such as SentencePiece Unigram and Byte Pair Encoding could be combined with the grapheme-aware preprocessing strategy to study whether the observed improvements arise primarily from grapheme preservation or from the interaction between grapheme-aware preprocessing and WordPiece vocabulary learning.


\section{Conclusion}

This paper presented a grapheme-aware Indic tokenizer for Tamil together with a comprehensive intrinsic evaluation of its tokenization quality. The proposed approach preserves complete grapheme clusters through a reversible Unicode mapping strategy before WordPiece vocabulary learning, enabling vocabulary construction over linguistically meaningful units while remaining fully compatible with existing transformer tokenization pipelines.

The tokenizer was trained on a large-scale Tamil corpus, and the influence
of vocabulary scaling was systematically investigated by training models
with vocabulary sizes ranging from 10K to 50K. Experimental results
demonstrate that increasing vocabulary size consistently improves
intrinsic segmentation quality by reducing token fragmentation,
increasing the amount of information represented by each token, and
producing shorter token sequences. At the same time, vocabulary coverage
remains essentially constant, indicating that vocabulary expansion
primarily enhances segmentation efficiency rather than lexical coverage.

Comparative evaluation against widely used multilingual tokenizers
including GPT-2, mBERT, mT5, mBART, and NLLB further demonstrates the
effectiveness of the proposed tokenizer within the evaluated setting.
The Grapheme-Aware Indic Tokenizer achieves the lowest Fertility, Word
Fragmentation Rate, Percentage of Pretokens Changed, and Normalized
Sequence Length while simultaneously achieving the highest Characters
per Token among the evaluated models. Despite using a language-specific
vocabulary, it also matches the highest observed Compression Ratio,
indicating efficient tokenization for Tamil.

Overall, the results demonstrate that preserving grapheme boundaries during vocabulary learning substantially improves intrinsic tokenization quality for Tamil. The proposed grapheme-aware preprocessing strategy provides a simple yet effective way to incorporate script-specific linguistic structure into modern subword tokenization while remaining compatible with existing transformer architectures. The methodology and evaluation framework presented in this work establish a foundation for extending grapheme-aware tokenization to additional Indic languages and for investigating its impact on downstream language modelling tasks.


\nocite{*}  
\bibliographystyle{IEEEtran}
\bibliography{references}

\end{document}